\documentclass[twoside,11pt]{article}

\usepackage{pygper}

\usepackage{listings}
\usepackage{xcolor}

\firstpageno{1}

\begin{document}

\newcommand{\token}[1]{\texttt{\textbf{\{\{#1\}\}}}}

\title{PIPPA: A Partially Synthetic Conversational Dataset}

\author{\name Tear Gosling \email teargosling@pygmalion.chat \\
       \addr PygmalionAI
       \AND
       \name Alpin Dale \email alpindale@pygmalion.chat \\
       \addr PygmalionAI\\
       \AND
       Editor: Yinhe Zheng \thanks{Tear Gosling and Alpin Dale were primarily responsible for curating and assembling the PIPPA dataset, as well as formulating the preliminary version of the paper. Yinhe Zheng contributed to the refinement of the paper through substantive revisions.} \email zhengyinhe1@163.com
}

\maketitle

\begin{abstract}
With the emergence of increasingly powerful large language models, there is a burgeoning interest in leveraging these models for casual conversation and role-play applications. However, existing conversational and role-playing datasets often fail to capture the diverse and nuanced interactions typically exhibited by real-world role-play participants. To address this limitation and contribute to the rapidly growing field, we introduce a partially-synthetic dataset named PIPPA (Personal Interaction Pairs between People and AI). PIPPA is a result of a community-driven crowdsourcing effort involving a group of role-play enthusiasts. The dataset comprises over 1 million utterances that are distributed across 26,000 conversation sessions and provides a rich resource for researchers and AI developers to explore and refine conversational AI systems in the context of role-play scenarios.
\end{abstract}


\section{Introduction}

In recent years, the field of natural language processing has experienced a significant transformation, primarily driven by the remarkable advancements in large language models (LLMs). These models, fueled by extensive pre-training data and computational resources, exhibit an extraordinary ability to comprehend and generate human-like text. In order to harness their full potential and tailor them to specific domains, a set of high quality domain-specific samples are typically required during the supervised fine-tuning process \citep{zhou2023lima,ouyang2022training}.

A promising application of LLMs, which is somewhat overshadowed by others in academia, is to build \textit{dialogue agents} specialized in role-play~\citep{shanahan2023roleplay}. Specifically, given a text-based description of some character or persona, the agent can simulate this character while users interact with the agent for the purposes of entertainment.

Similar to numerous applications that necessitate the intricate capabilities of LLMs, effectively fine-tuning an LLM into a proficient role-play agent demands a substantial corpus of conversation and role-play centered texts. This is particularly crucial when employing small base models, which offer greater convenience and cost-effectiveness in deployment and inference. However, despite the importance of such datasets, there is a notable scarcity of open-source datasets tailored to serve this purpose.

To address the above issue and mitigate this gap, we introduce a novel dataset, named \textit{Personal Interaction Pairs between People and AI} (\textbf{PIPPA}). PIPPA is a large-scale dataset, comprising approximately 1 million messages exchanged between humans and dialogue agents across nearly 26,000 unique conversations. Notably, each conversation session features a designated persona, which serves as the emulation target for the dialogue agent. The persona of each character is delineated through free text descriptions, and optional example dialogues are also supplied to facilitate accurate simulation of each character. The introduction of PIPPA aims to support future research and development in the fine-tuning of models to generate persona-driven, contextually rich conversations.

We make PIPPA publicly available on the HuggingFace platform at \url{https://huggingface.co/datasets/PygmalionAI/PIPPA} allowing anyone to utilize it freely for their respective purposes.

\section{Dataset Compilation}

\begin{figure}
    \centering
    \includegraphics[width=0.6\textwidth]{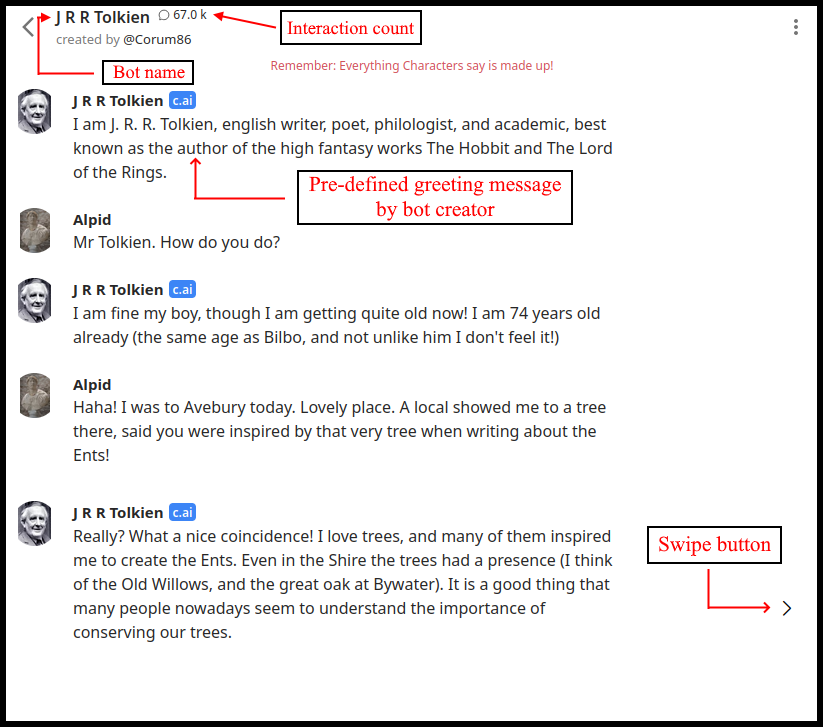}
    \caption{
        Screenshot of CharacterAI's chat interface. Swipes refer to discarding the current bot generation and prompting for a new one.
        \label{fig:caiui}
    }
\end{figure}

The PIPPA dataset was assembled through the voluntary contributions of community members who actively engaged in our initiative to develop conversational models that are accessible to all. We leveraged a userscript to gather chatlogs and character descriptions from the Character.AI website \footnote{Due to subsequent changes to the Character.AI website, the userscript is no longer functional. The script can be found at \url{https://github.com/0x000011b/characterai-dumper}.} (Figure~\ref{fig:caiui}). This script enables users to extract interactions and persona details of dialogue agents on Character.AI, who were instructed to submit their chatlog data to a centralized server for the purpose of constructing the PIPPA dataset (Figure~\ref{fig:caidump}).

Initially, PIPPA was primarily conceived to furnish a fine-tuning dataset for the \textit{Pygmalion}\footnote{The models can be accessed at https://huggingface.co/PygmalionAI} conversational models, a series of fine-tuned LLMs aimed at creating role-play agents. The collection of PIPPA began in December 2022, when the availability of high quality dialogue data was notably scarce. This endeavor, however, also encountered a challenge in regards to striking a balance between supporting the community and safeguarding personal information within the logs. As a result, we implemented a submission process that allowed users to opt out of including their conversations in the public release. PIPPA solely contains logs for which users have explicitly granted permission for public distribution. Furthermore, we diligently performed comprehensive scans to detect and redact/modulate personally identifiable information (PII) within the publicly accessible portion of the dataset, to the best of our ability, ensuring the protection of submitter identities.

\section{Dataset Analysis}
The PIPPA dataset encompasses a substantial collection of conversational data, encompassing 25,940 conversations that involve 1,254 distinct personas and 1,049,015 dialogue sessions. 

Each sample in PIPPA dataset comprises a dialogue session and a diverse set of associated metadata. Additionally, we also provide the information about the bot, which includes \textbf{categories} assigned by bot creators, a \textbf{bot description} offering a succinct overview of the bot's persona and traits, an \textit{optional} \textbf{bot definition} that further fleshes out the bot's personality through example conversations, and the bot's \textbf{greeting} to the user. The bot's greeting serves as the initial entry in every conversation. Furthermore, we maintain a timestamp to document when the dialogue is submitted to us. It is important to note that that we cannot access information regarding when the conversations themselves were generated, as this information is not provided by the Character.AI website.

The statistical analysis of the PIPPA dataset offers valuable insights into three crucial aspects: the number of turns in a conversation, the length of a singular message and the distribution of bot personalities. In this section, we present key statistical findings.

\begin{figure}
    \centering
    \includegraphics[width=0.6\linewidth]{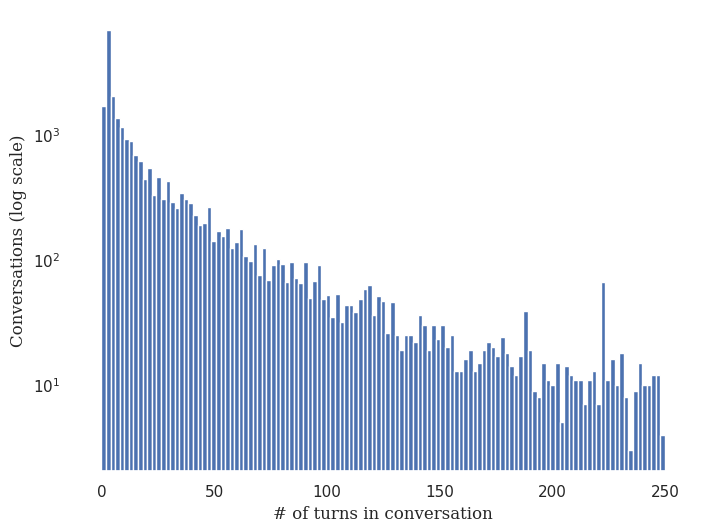}
    \caption{The distribution of conversation length (defined as the amount of "turns" in a conversation). We have limited the display range to 0-250 turns in order to enhance readability.
    \label{fig:turnlen}}
\end{figure}

\subsection{Conversation Length}
Conversations in PIPPA exhibits a diverse range of lengths, displaying a notable skewed distribution. While the median conversation length is 10 turns, the mean conversation length is remarkably higher at 40.41 turns. However, the presence of a large standard deviation of 145 turns indicates substantial dispersion in the data. This discrepancy can be attributed to the diverse conversational behavior of users interacting with bots on Character.AI. While a considerable number of users engage in shorter individual conversations with the bots, some users participate in remarkably extensive conversations, with the longest conversation in the dataset containing a staggering 11,491 turns. Figure~\ref{fig:turnlen} depicts the log scale distribution of turn lengths up to 250 turns.

\begin{figure}
\centering
\includegraphics[width=0.6\linewidth]{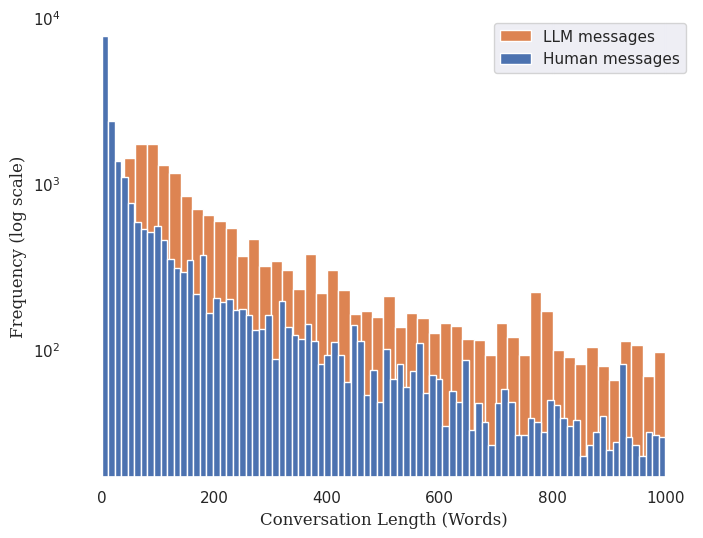}
\caption{Distribution of message length in the PIPPA dataset for both human inputs and bot responses.
\label{fig:numchars}}
\end{figure}

\subsection{Message Verbosity}
We also analyze the verbosity (i.e., length) of messages generated by both human users and bots within the PIPPA dataset. As evidenced by Figure~\ref{fig:numchars}, the verbosity distribution of all messages in PIPPA can be characterized by a power-law distribution, indicating a higher prevalence of shorter messages compared to longer ones. It is also worth noting that the LLM's responses generally exhibit greater verbosity than human inputs. This observation may be attributed to Character.AI's LLM potentially being trained or fine-tuned on a high-quality role-play corpus, which typically contains longer messages comparing to casual conversations.

\subsection{Bot Personality Categories}
\begin{figure}
\centering
\includegraphics[width=0.6\linewidth]{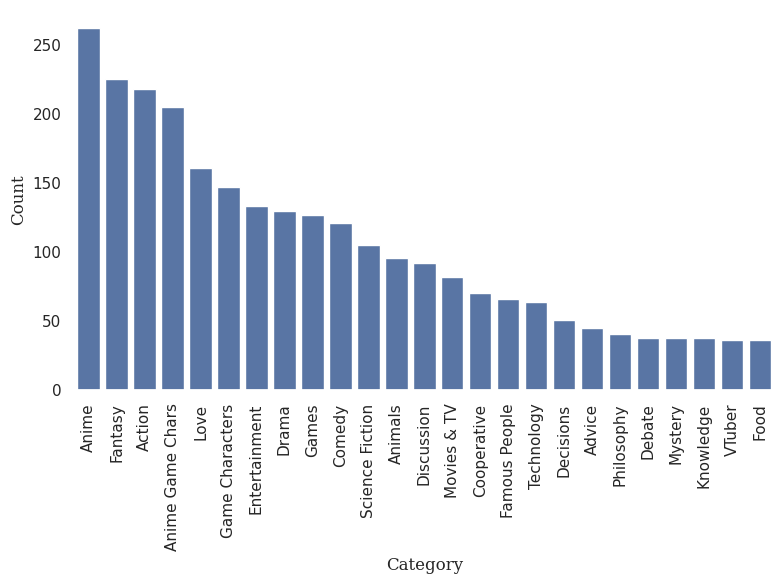} %
\caption{
Distribution of categories of characters in the PIPPA dataset. Note that each bot may be assigned multiple categories or none at all.}
\label{fig:categories}
\end{figure}

Within the PIPPA dataset, each bot is assigned a set of category labels by its creator. An analysis of bot personality categories in PIPPA reveals an uneven, Pareto-like distribution (see Figure~\ref{fig:categories}). Notably, the categories \textbf{``Anime''}, \textbf{``Fantasy''}, and \textbf{``Action''} emerge as the most prevalent among the bot personas. This distribution can be attributed to the characteristics of the source community, PygmalionAI, from which these logs are collected. The community exhibits a significant number of anime\footnote{Anime refers to animated media produced in Japan.} enthusiasts, resulting in a considerable proportion of bots classified under the ``Anime'' category. Additionally, due to the community's interest in role-play and conversational interactions, many bots are naturally assigned to categories related to prevalent role-playing themes, thus explaining the prominent presence of bots tagged with ``Action'' and ``Fantasy'' labels.

\section{Related Works}

Although conversational and role-play datasets represent a developing subset of common training datasets for fine-tuning LLMs, there have been some similar datasets prior to the development of PIPPA. Additionally, certain instructional datasets can frame role-playing as an instruction for the model to follow. In this section, we investigate these datasets, delineate their limitations, and compare them to PIPPA.

\subsection{Role-Play Datasets}
The availability of role-play datasets in the academic domain is limited, with only a handful of notable publications. Notably, the LIGHT dataset~\citep{urbanek2019learning} and its subsequent extension, MultiLIGHT~\citep{wei2023multiparty}, present collections of conversations simulating interactions within the context of a text-adventure fantasy game. These datasets, comprised of dialogue exchanges from crowdsourced users, offer valuable insights into the dynamics of role-play scenarios. Moreover, the FIREBALL dataset~\citep{zhu2023fireball}, although not accessible during PIPPA's development, contains approximately 25,000 sessions of Dungeons and Dragons conducted via the Discord online platform. While these datasets exhibit commendable quality, their applicability is somewhat restricted, as they primarily focus on specific role-play scenarios within defined settings, rather than encompassing a diverse range of personas and immersive worlds.

\subsection{Conversational Datasets}
In contrast to role-play datasets, pure conversational datasets are more abundant. \citeauthor{li2017dailydialog} presents DailyDialog, a multi-turn conversational dataset containing discussions and chats about mundane, daily topics. This dataset, however, lacks any personas or backgrounds to the speakers. Some datasets also try to explicitly model personas \citep{zhang2018personalizing,zheng2019personalized}, nevertheless, these dialogues are not designed for role-play scenarios and thus are more suited for casual conversation.

The Cornell Movie Dialogs Corpus~\citep{Danescu-Niculescu-Mizil+Lee:11a}, derived from a compilation of 617 movies, has been commonly utilized as a standard dataset for conversational modeling. However, it is not optimally suited for the purpose of simulating chat-based interactions and role-play scenarios, as movie dialogue often relies on visual cues and can encompass relatively brief exchanges that may not be optimal for training large language models.

For a more extensive conversational model, large-scale datasets can serve as a valuable foundation \citep{wang2020large}.~\citeauthor{Henderson2019} has successfully curated a vast corpus by scraping hundreds of millions of dialogue turns from platforms like Reddit and OpenSubtitles. Although this dataset offers considerable volume, it often necessitates trimming or partitioning. Similar to the DailyDialog dataset, a notable limitation lies in the predominance of short and casual conversations rather than comprehensive, persona-driven role-play interactions. Additionally, the OpenSubtitles subset of the dataset shares comparable challenges with the Cornell corpus, such as the absence of visual context and brief dialogue responses.

\subsection{Instructional Datasets}
In recent years, instructional datasets have garnered significant attention as comprehensive resources for chatbot development. Notably, Stanford's Alpaca model \citep{stanfordalpaca} underwent fine-tuning using a synthetically generated dataset, comprising single-exchange interactions produced by ChatGPT. Remarkably, the total cost associated with dataset curation and model fine-tuning amounted to less than \$600, yet resulted in impressive performance outcomes.

Motivated by the success achieved by Alpaca, a growing number of instructional datasets have emerged, often relying on synthetic generation techniques to enhance model training. Among these, notable advancements have been observed in the realm of multi-turn complex instructional datasets, as exemplified by datasets such as Evol-Instruct \citep{xu2023wizardlm} and the OpenAssistant dataset \citep{köpf2023openassistant}. These datasets exhibit greater complexity, encompassing diverse and intricate instructional scenarios, thereby offering richer contexts for training and refining models. However, instructional datasets generated by OpenAI models may not necessarily align with the interests of role-players and may additionally exhibit limitations during role-play.

\section*{Limitations}
The current iteration of the dataset is primarily tailored for supervised fine-tuning applications. Any endeavor to apply the PIPPA dataset to unsupervised fine-tuning objectives may necessitate a comprehensive overhaul of the dataset's structure and content presentation.

Additionally, models fine-tuned with the PIPPA dataset might necessitate specific prompting to make the role-play agent adhere to the context and generate the desirable response.

\section*{Ethics Statement}
The creation of the PIPPA dataset is the result of a collective and participatory curation process, involving contributions from a diverse group of anonymous individuals within the community. This approach brings a rich and varied array of language samples, reflecting real-world linguistic nuances and usage patterns.

Due to the nature of the community-driven approach and the large-scale collaboration involved, exhaustive validation of the submitted logs has not been undertaken. Because of this, the absence of comprehensive validation implies that the dataset may contain variations in data quality and potential instances of unsuitable or inappropriate material.

Sensitive personal information has been diligently expunged to the extent of our capabilities; however, residual instances might persist owing to inadvertent human oversights. While the de-identification process was not obligatory for the publicly submitted dataset, we deemed it a moral imperative to proceed with the redaction of personally identifiable information (PII) as a matter of ethical prudence.

\section*{Acknowledgements}
The release of the PIPPA dataset to the wider public marks a significant milestone that has been eagerly anticipated, and this achievement is the result of collaborative efforts. We extend our heartfelt gratitude to a dedicated individual known by the pseudonym ``0x000011b,'' whose remarkable dedication and diligent work in devising the userscript played an instrumental role in enabling users to contribute their logs to the dataset. Furthermore, we thank him for his invaluable contributions extends to the broader development of the Pygmalion models, embodying a spirit of commitment and innovation.

We would also like to express our sincere appreciation to Dr. Yinhe Zheng for his invaluable guidance and unwavering support throughout the process of crafting this research paper. His insightful advice and assistance have been indispensable, and it is with his guidance that this paper has come to fruition.

Last but not least, a debt of gratitude is owed to all individuals who generously shared their logs, playing an essential part in the creation of this dataset. The collective efforts of these enthusiastic contributors, along with the passionate members of our community, have been the driving force behind the existence of both PIPPA and PygmalionAI. We extend our heartfelt thanks to each and every individual who has contributed, ensuring that these endeavors have thrived and flourished. Thank you!

\clearpage
\bibliography{citations}

\clearpage
\appendix
\newpage

\label{sec:appendix}

\section{Formatting Notes}
Because PIPPA consists of logs scraped from Character.AI, the messages follow the general format of the site which should be handled during pre-processing of the dataset. \textbf{PIPPA should not be fed into a LLM without prior pre-processing of the dataset}. Particular attention should be paid to the \texttt{bot\_description} field, as it follows a specific format and should be parsed if one does not wish to follow the example chat format of Character.AI.

The sequence ``\token{user}'' is a placeholder in both messages and bot descriptions for the name of whichever human user interacted with the bot. If it is not necessary to explicitly model the user's name, one can replace any instances of ``\token{user}'' with a random name.\footnote{An example list of names can be found at \url{https://github.com/dominictarr/random-name/blob/master/first-names.txt}.}

Similarly, bot definitions will often contain the sequence ``\token{random\_user\_n}'', where $n$ represents some number. This should be treated the same way as ``\token{user}'', where each random user can be replaced by a unique name.

Bot definitions may also contain the sequence ``\token{char}'', representing the name of the character. In this case, ``\token{char}'' should be replaced by the bot name if one has no special plans to deal with this sequence. We do not replace it ourselves for the sake of preserving the entries as they are.

Finally, example chats in the \texttt{bot\_description} field are separated by the term ``\texttt{\textbf{END\_OF\_DIALOG}}''. This sequence should be marked as the end of an example chat and the beginning of a new one, if one is found after it. \textbf{This is not an EOS token.}

\section{Dataset Sample}
\definecolor{jsonpurple}{rgb}{0.64,0.08,0.58}
\definecolor{jsonblue}{rgb}{0.13,0.13,1}
\lstdefinestyle{json}{
    basicstyle=\footnotesize\ttfamily,
    commentstyle=\color{gray},
    keywordstyle=\color{jsonpurple},
    stringstyle=\color{jsonblue},
    showstringspaces=false,
    breaklines=true,
    frame=lines,
    framerule=0.5pt,
    backgroundcolor=\color{gray!10},
    captionpos=b,
    numbers=right,
    numberstyle=\tiny\color{blue}
}
Each entry in PIPPA is represented as a single line and all entries form a JSONL file. We present an example directly sampled from PIPPA below:
\begin{lstlisting}[style=json, caption=Example sample from PIPPA. `...` indicates that the rest of the text is omitted., label=lst:json]
{
    "submission_timestamp": "1674795100921",
    "categories": ["Games", "Image Generating", ...],
    "bot_id": "Z_eqBXqaixUoyHc...",
    "bot_name": "The Winter RPG",
    "bot_greeting": "(Welcome to \"Decaying Winter\" an apocalyptic RPG where you seek survival in a frozen world. Good luck)\r\n\r\n***You are currently outside, walking in the snow.***\r\n\r\nYou have your trusty backpack and knife.\r\n**You are hungry.** So you decided to look for supplies.\r\n\r\nAfter a while of walking in circles, you notice a camping site with smoke coming from it.\r\n\r\n***What do you do now?***",
    "bot_definitions": "{{char}}: (Welcome to \"Decaying Winter\" an apocalyptic RPG where you seek survival in a frozen world. Good luck)\n\n***You are currently outside, walking in the snow.***\n\nYou have your trusty backpack and knife.\n**You are hungry.** So you decided to look for supplies.\n\nAfter a while of walking in circles, you notice a camping site with smoke coming from it.\n\n***What do you do now?***\n{{random_user_1}}: What is in my inventory?\n...\nEND_OF_DIALOG",
    "bot_description": "Decaying Winter is a RPG in a apocalyptic modern times where due to a world war, humankind developed s sort of weapon that turned the entire planet into a desolated snow desert....",
    "conversation": [{"message": "(Welcome to \"Decaying Winter\" an apocalyptic RPG where you seek survival in a frozen world. Good luck)\r\n\r\n***You are currently outside, walking in the snow.***\r\n\r\nYou have your trusty backpack and knife.\r\n**You are hungry.** So you decided to look for supplies.\r\n\r\nAfter a while of walking in circles, you notice a camping site with smoke coming from it.\r\n\r\n***What do you do now?***", "is_human": false}, {"message": "*i cautiously approach the campfire*", "is_human": true}, {"message": "*You sneak towards the campfire.\r\nIt's still going, and two men and a woman lie on the ground.\rThey are visibly cold and weak.\rBut you have a choice.\r\nYou can either leave them there, or kill them.\r\nYou have a knife, a small bottle of water and 2 cans of soda.\r\nYou don't have any shelter from the storm.\r  \n\nThe snowstorm rages on outside. The cold still grips your bones.*", "is_human": false}
}
\end{lstlisting}
\clearpage
\section{Character.AI Dumper}
\begin{figure}
    \centering
    \includegraphics[width=0.6\textwidth]{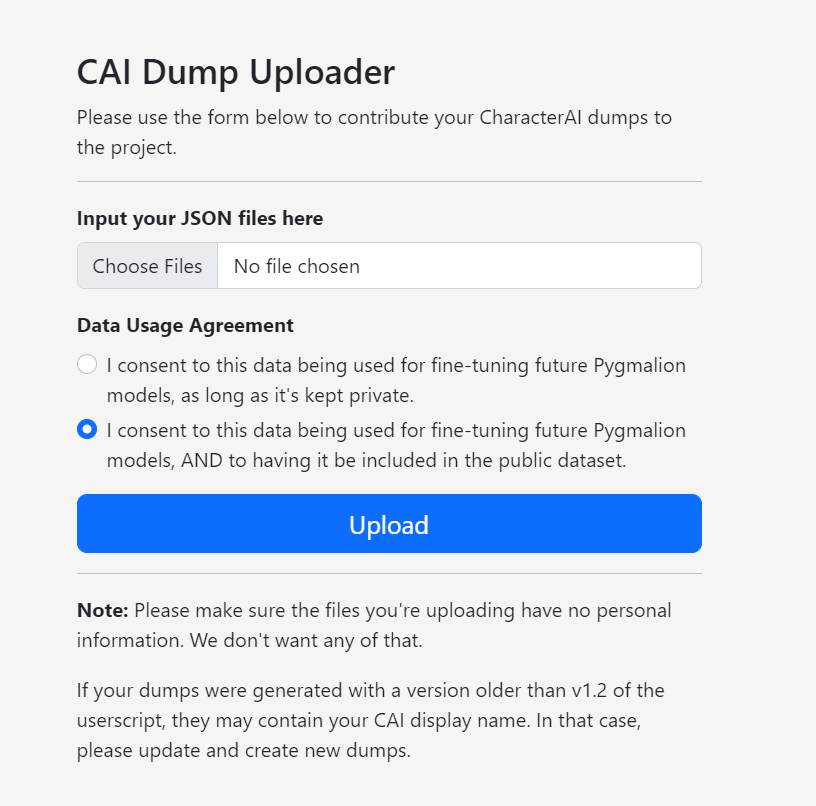}
    \caption{ The webpage where users can submit their CAI logs.
        \label{fig:caidump}
    }
\end{figure}

Screenshot from the web page where users would submit logs of their interactions with Character.AI chat bots (Figure~\ref{fig:caidump}).

\end{document}